\documentclass[letterpaper, 10 pt, conference]{ieeeconf}  

\IEEEoverridecommandlockouts                              

\usepackage{graphicx}
\usepackage{multirow}
\usepackage{float}
\graphicspath{ {./images/} }
\usepackage{multicol, blindtext}
\usepackage{cite}

\title{\LARGE \bf
Prediction of Metacarpophalangeal joint angles and Classification of Hand configurations based on Ultrasound Imaging of the Forearm
}

\author{Keshav Bimbraw$^{1}$, Christopher J. Nycz$^{2}$, Matt Schueler$^{1}$, Ziming Zhang$^{3}$ and Haichong K. Zhang$^{1}$$^{*}$
\thanks{*Corresponding author}
\thanks{$^{1}$Keshav Bimbraw, Matthew Schueler and *Haichong K. Zhang are with the Medical FUSION Lab at Worcester Polytechnic Institute, 100 Institute Rd, Worcester, MA 01609, USA.
        {\tt\small hzhang10@wpi.edu}}%
\thanks{$^{2}$Christopher Julius Nycz is a research scientist working at PracticePoint, Worcester Polytechnic Institute, 50 Prescott Street, Worcester, MA 01605, USA.
        {\tt\small cjnycz@wpi.edu}}%
\thanks{$^{3}$Ziming Zhang is with the Vision, Intelligence, and System Laboratory (VISLab) at Worcester Polytechnic Institute (WPI), 100 Institute Rd, Worcester, MA 01609, USA.
        {\tt\small zzhang15@wpi.edu}}%
}

\begin{document}

\maketitle
\thispagestyle{empty}
\pagestyle{empty}

\begin{abstract}

With the advancement in computing and robotics, it is necessary to develop fluent and intuitive methods for interacting with digital systems, augmented/virtual reality (AR/VR) interfaces, and physical robotic systems. Hand movement recognition is widely used to enable this interaction. Hand configuration classification and Metacarpophalangeal (MCP) joint angle detection are important for a comprehensive reconstruction of the hand motion. Surface electromyography (sEMG) and other technologies have been used for the detection of the hand motions. Ultrasound images of the forearm offer a way to visualize the internal physiology of the hand from a musculoskeletal perspective. Recent work has shown that these images can be classified using machine learning to predict various hand configurations. In this paper, we propose a Convolutional Neural Network (CNN) based deep learning pipeline for predicting the MCP joint angles. We supplement our results by using a Support Vector Classifier (SVC) to classify the ultrasound information into several predefined hand configurations based on activities of daily living (ADL). Ultrasound data from the forearm was obtained from 6 subjects who were instructed to move their hands according to predefined hand configurations relevant to ADLs. Motion capture data was acquired as the ground truth for hand movements at different speeds (0.5 Hz, 1 Hz, \& 2 Hz) for the index, middle, ring, and pinky fingers. We were able to get promising SVC classification results on a subset of our collected data set. We demonstrated a correspondence between the predicted MCP joint angles and the actual MCP joint angles for the fingers, with an average root mean square error of 7.35 degrees. We implemented a low latency (6.25 - 9.1 Hz) pipeline for the prediction of both MCP joint angles and hand configuration estimation aimed at real-time control of digital devices, AR/VR interfaces, and physical robots.


\end{abstract}

\section{Introduction}

Smart and intuitive interaction with physical and non-physical worlds (AR/VR) is a topic of great interest to the human-computer interaction research community and has numerous interfacing and control applications. Upper limb motor dexterity in humans is possible with a highly advanced neuromuscular machinery developed over millions of years of evolution from the primitive primates with prehensile appendages to our current upper limbs. For humans, our upper limbs have the maximum proportion of motor and sensory innervation in the human body and we use them to manipulate and interact with the physical world. There is a growing interest in developing technologies that can facilitate a fluent and intuitive upper limb interaction and interfacing with digital, AR/VR and remote domains such as teleoperated robots \cite{tarasenko2020artificial}. Several physical sensors and vision based techniques have been utilized to obtain the MCP joint angles which can then be reconstructed in digital, physical and AR/VR environments \cite{cheok2019review}. Placing sensors on the fingers, such as bend sensors and motion sensors can limit mobility and usage of the hands to their full extent for the individuals wearing such sensors. Vision based methods for hand state and orientation detection are dependent on the light condition and line of sight. Therefore, biological signals from the forearms can be used as a better alternative to understanding hand movement without impeding the user and preventing difficult and awkward hand movements.
\begin{figure*}[t]
\centering
\includegraphics[scale=0.62]{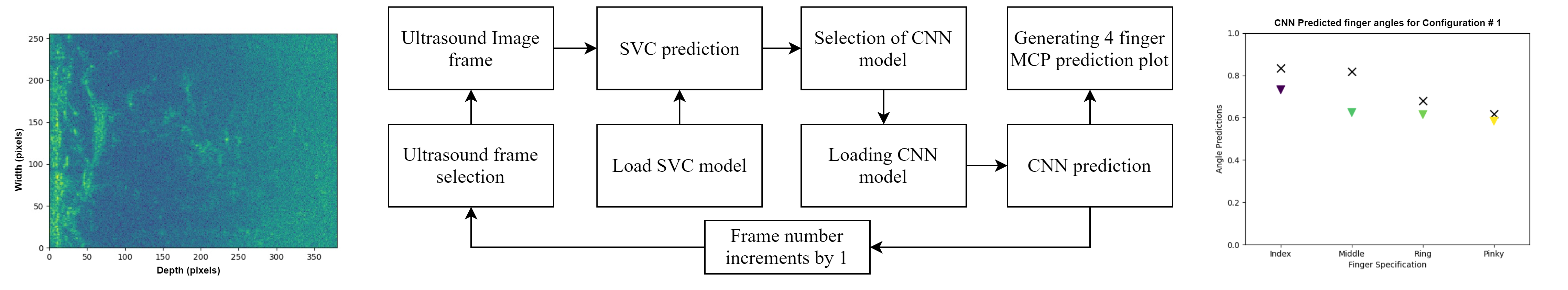}
\caption{The combined pipeline for hand configuration classification and MCP joint angle prediction. It shows a portion of the input reconstructed ultrasound image on the left and output prediction plot for the combined system demo. The crosses represent the actual data and the colored inverted triangles represent the predicted data for the four fingers.}
\label{fig:proces}
\end{figure*}
\subsection{Related Works}
Various contact based and non-contact based technologies have been used to understand hand motion. These include sEMG \cite{ahsan2009emg}, FMG \cite{jiang2017exploration}, vision based approaches \cite{oudah2020hand}, resistive hand gloves \cite{guo2021human}, depth information based approaches \cite{suarez2012hand} and WiFi sensing \cite{ahmed2020device}. sEMG has been primarily widely used for understanding and decoding the hand movements \cite{li2020review}. It has been researched for its use as a control input to prosthetics and powered orthotics \cite{graupe1982multifunctional}. Various combinations of sensors with sEMG has also been explored for upper and lower limb motion prediction \cite{lee2020image, kahanowich2021robust, dong2021soft, rabe2020use}.  Araki et al. used sEMG to measure middle finger joint angle and used it to control a robotic finger \cite{araki2012artificial}. Shrirao et al. used sEMG to predict index finger joint angle while performing flexion-extension rotation of the index finger \cite{shrirao2009neural}. Wang et al. used sEMG to predict joint angles during different grasps \cite{wang2020semg}.  Deep Learning has also been used for facilitating sEMG based human computer interaction \cite{xiong2021deep}. sEMG is sensitive to muscle fatigue as a consequence of long term muscle movements, and works best for fast movements of the hand which can inhibit human-like control of digital and virtual interfaces, and prosthetics/orthotics \cite{li2020review, shrirao2009neural}.

While sEMG can be used to give an estimate of the muscle activation for hand and finger movements, ultrasound imaging of the forearm can be used to visualize the muscles which can be used for hand state prediction using image processing techniques. Ultrasound imaging of the forearm, or Sonomyography, has been explored as an alternative sensing modality that can capture both muscle configuration and movement \cite{bimbraw2020towards,  zheng2006sonomyography, chen2010sonomyography}. It has been shown to be capable of identifying different hand gestures and finger movements by analyzing the image obtained from ultrasound data with a combination of image processing and classification algorithms \cite{bimbraw2020towards, shi2010feasibility}. Sonomyography has also been used to classify several grasp types and controlling robotic mechanisms \cite{bimbraw2020towards}. Akhlagi et al. demonstrated classification of various hand gestures using Nearest Neighbor classification algorithms \cite{akhlaghi2015real}. Wrist motion classification has also been done utilizing ultrasound imaging \cite{yang2020simultaneous}. McIntosh et al. classified 10 hand gestures based on small ultrasound data sizes using support vector machines (SVMs) and multi layer perceptrons (MLPs) \cite{mcintosh2017echoflex}.

\subsection{Contributions}
While hand gesture recognition has been implemented previously, there remains to be a need for continuous prediction of the MCP joint angles for potential usage as a direct or proportional control of systems and environments. With the advances in deep learning, newer algorithms and techniques can be used to make useful predictions from ultrasound images.  We can use this to predict finer hand movements primarily the MCP joint angles while attaining hand configurations relevant to ADL. In this paper, we present our work on utilizing a CNN for predicting MCP joint angles based on ultrasound imaging. We supplement our implementation by using an SVC for classifying different hand motions. We also include results from our simultaneous state/angle prediction pipeline. This work is the foundation of our future research in the domain of human-computer and human-robot interaction that can be facilitated by ultrasound imaging.

\section{Methods}
We used an SVC for hand state classification and a CNN for predicting finger angles based on the ultrasound images from the forearm. We also developed a low-latency demo to show the performance of our combined hand state classification and finger angle prediction pipeline.

\subsection{Hand Configuration Classification}
SVMs are a class of classical machine learning algorithms based on supervised learning which are used for image classification and regression problems. For ultrasound forearm image classification to predict hand configurations, SVC and other classical machine learning algorithms have been explored in the past \cite{bimbraw2020towards, mcintosh2017echoflex}. A support vector classifier constructs hyperplanes in high-dimensional spaces, wherein a good separation is achieved by the hyperplane that has the largest distance to the nearest training-data point of any class. 

\subsection{MCP joint angle estimation}
MCP joint angle prediction was implemented using a CNN. CNNs are a class of deep learning algorithms popularly used for two dimensional image data regression based prediction. The CNN architecture that was used in the paper is based on VGG16 \cite{simonyan2014very}.

\subsection{Combined System Demo}

The combined system demonstration pipeline uses a single script to use saved SVC and CNN models to predict both the hand configuration and MCP joint angles. First, the ultrasound image files are loaded and SVC model is loaded. The SVC model prediction decides the model to be loaded for the CNN. Once the CNN model is loaded, the CNN prediction is executed and the four finger MCP angle prediction plot is displayed. This is done repeatedly for a given number of frames which can be specified by the user as a way to demonstrating the combined pipeline shown in Fig \ref{fig:proces}.

\begin{figure*}[t]
\centering
\includegraphics[scale=0.6]{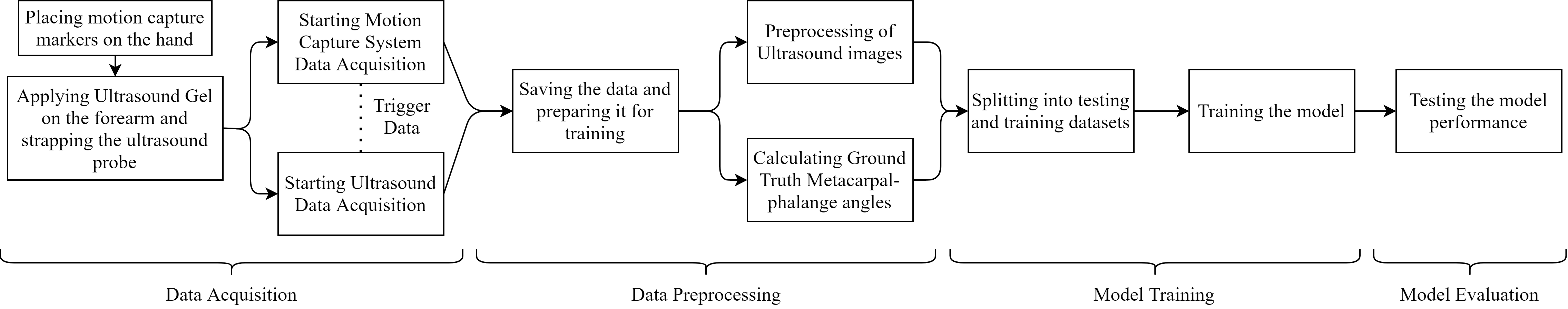}
\caption{The data acquisition, processing and evaluation pipeline. Primarily, there are four parts to the pipeline. First, ultrasound data and motion capture ground truth is collected. Then, data pre-processing is done to obtain ultrasound images and ground truth MCP joint angles. The data is then split to testing and training sets, and then the deep learning and machine learning models are trained using the training data and evaluated on the test data for MCP joint angle estimation and hand configuration classification.}

\label{fig:process1}
\end{figure*}

\begin{figure}[h]
\centering
\includegraphics[scale=0.35]{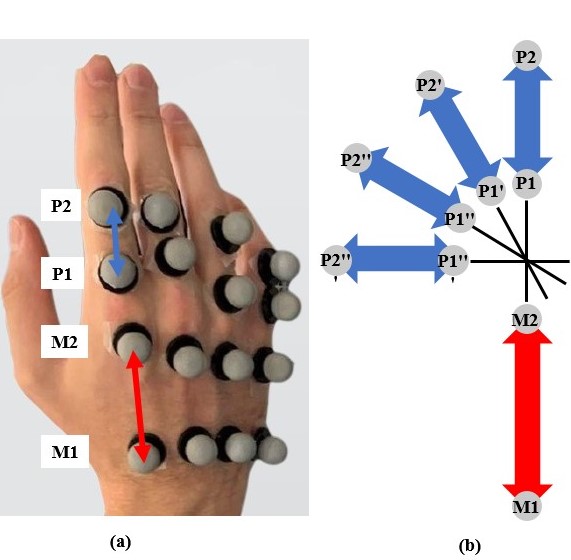}
\caption{(a) Location of markers and the vectors on the subject's hand. (b) The angle between the $\overrightarrow{M1 M2}$ and $\overrightarrow{P1 P2}$ is 180$^{\circ}$ measured counterclockwise. When the proximal phalanx flexes by 30$^{\circ}$ from its initial state ($\overrightarrow{P1 P2}$), the angle between the $\overrightarrow{M1 M2}$ and $\overrightarrow{P1' P2'}$ is 210$^{\circ}$. Similarly, when the proximal phalanx flexes by 30$^{\circ}$ and 60$^{\circ}$ from its initial state ($\overrightarrow{P1' P2'}$), the angle between the $\overrightarrow{M1 M2}$ and $\overrightarrow{P1'' P2''}$, and $\overrightarrow{M1 M2}$ and $\overrightarrow{P1''' P2'''}$ is 240$^{\circ}$ and 270$^{\circ}$.}
\label{fig:Hand_Angle}
\end{figure}

\section{Experimental Implementation}
Ultrasound data was captured from the right forearm of 6 subjects (1 female, 5 males; Age: 22.83 $\pm$ 3.14 years; Height: 173.55 $\pm$ 11.02 cm; Probe placement forearm diameter: 20.43 $\pm$ 2.74 cm). Three dimensional positional data of specified points on their index, middle, ring and pinky fingers was acquired along with the ultrasound data, with the ultrasound probe mounted in a transverse position on the forearm. This data was acquired at three different speeds of hand movement. 

The experimental setup uses a Vantage 128 Verasonics ultrasound data acquisition system (Verasonics, WA, USA) and a Vicon Nexus motion capture system (Vicon Motion Systems Ltd., UK). The detailed data acquisition and processing pipeline is shown in Fig \ref{fig:process1}.

\subsection{Human Subjects and IRB Approval}
The study was approved by the institutional research ethics committee at the Worcester Polytechnic Institute (No. IRB-21-0452), and written informed consent was given by the subjects prior to all sessions. Table \ref{tab:my_label} lists the age, sex, height and forearm diameter at the ultrasound probe location for the 6 subjects enrolled in the study.

\begin{table}[b]
\centering
\caption{Subject Information}
\label{tab:my_label}
\begin{tabular}{|p{1.2cm}||p{0.58cm}|p{0.58cm}|p{0.58cm}|p{0.58cm}|p{0.58cm}|p{0.58cm}|}
\hline
& S. \#1 & S. \#2 & S. \#3 & S. \#4 & S. \#5 & S. \#6\\
\hline
Age (yr) & 25 & 20 & 24 & 23 & 24 & 21\\
Sex (M/F) & F  & M & M & M & M & M\\
Height (cm) & 157.5 & 170.2 & 172.7 & 180.3 & 167.6 & 193.0\\
Forearm Dia. (cm) & 22.9 & 18.4 & 18.4 & 19.1 & 18.4 & 25.4\\
\hline
\end{tabular}
\end{table}

\subsection{Instrumentation}
\begin{figure}[t]
\centering
\includegraphics[scale=0.1]{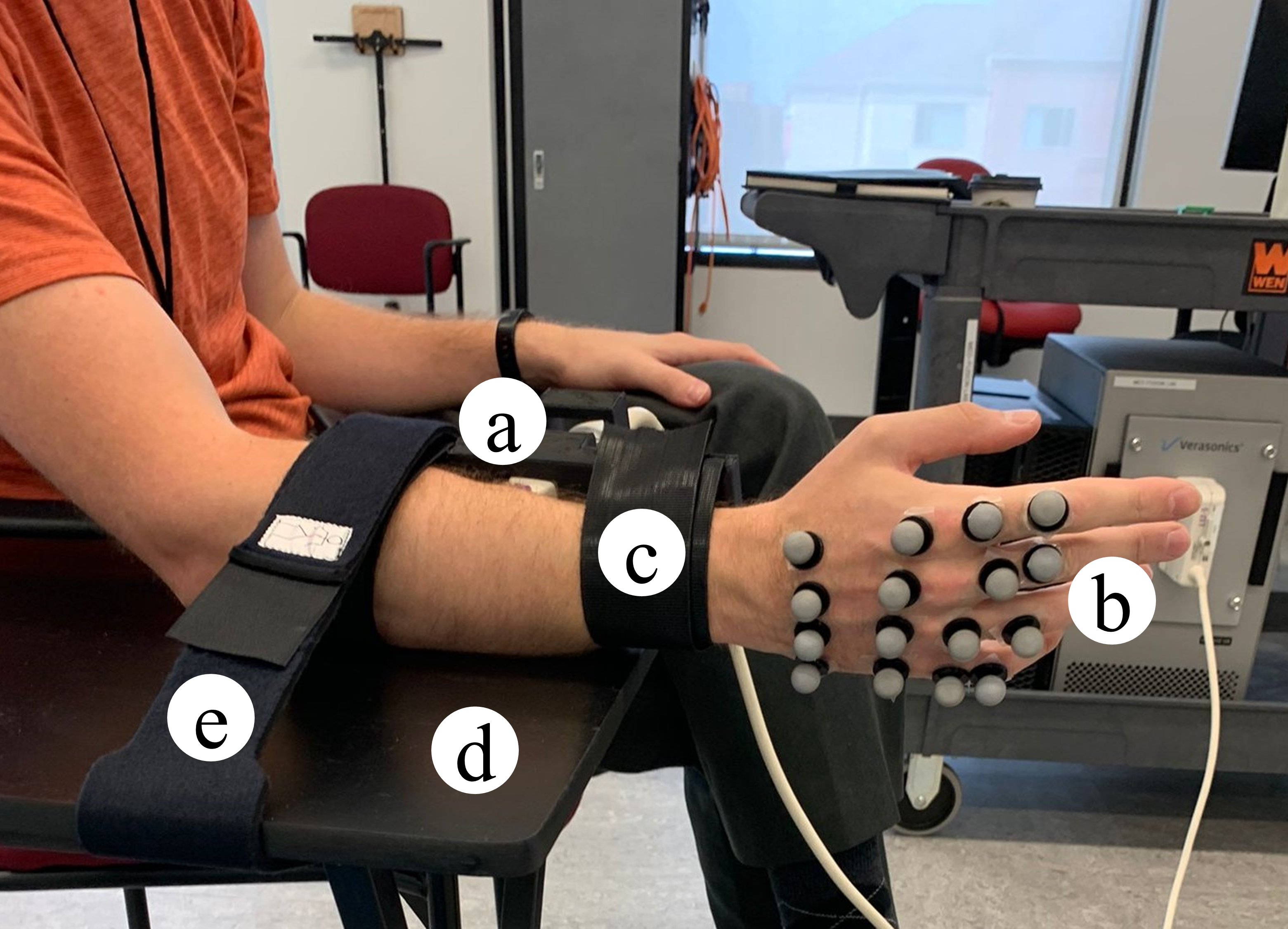}
\caption{The ultrasound probe in a 3D printed casing (a) strapped on a subject's right forearm (c) and the motion capture markers (b) on their right hand. The forearm is secured with a Velcro strap (e) around the rest table (d). The ultrasound probe is mounted transversally on the forearm.}
\label{fig:data_setup}
\end{figure}

Velcro straps on both sides of the custom designed 3D printed probe casing were used to strap the ultrasound probe on the forearm. The subject's arm rests on a rest table which was further secured with another external Velcro extender to the table after verification of the ultrasound images. Figure \ref{fig:data_setup} shows a subject's right arm strapped to the rest table.
\subsubsection{Verasonics Ultrasound System}
An L12/5 50 mm linear array ultrasound probe was used with the Verasonics Vantage 128/128 research ultrasound system. A MATLAB script was used to set sequence objects for the Verasonics data acquisition hardware to display and record ultrasound images. A trigger output signal was sent to an Arduino board from the Verasonics system. The ultrasound data frame acquisition rate was set at 25 Hz. For 56 seconds of data acquisition per session, 1400 frames were recorded. Each data frame was 636 x 256 pixels.
\subsubsection{Vicon Motion Capture System}
A 10 Camera Vicon Vantage motion capture system with Lock+ 64-channel ADC (Analog to Digital Converter) was used to get the three dimension positional information for different motion capture markers attached to the fingers. The system is present in the Motion Capture Suite in Worcester Polytechnic Institute's PracticePoint facility.
\begin{figure}[t]
\centering
\includegraphics[scale=0.24]{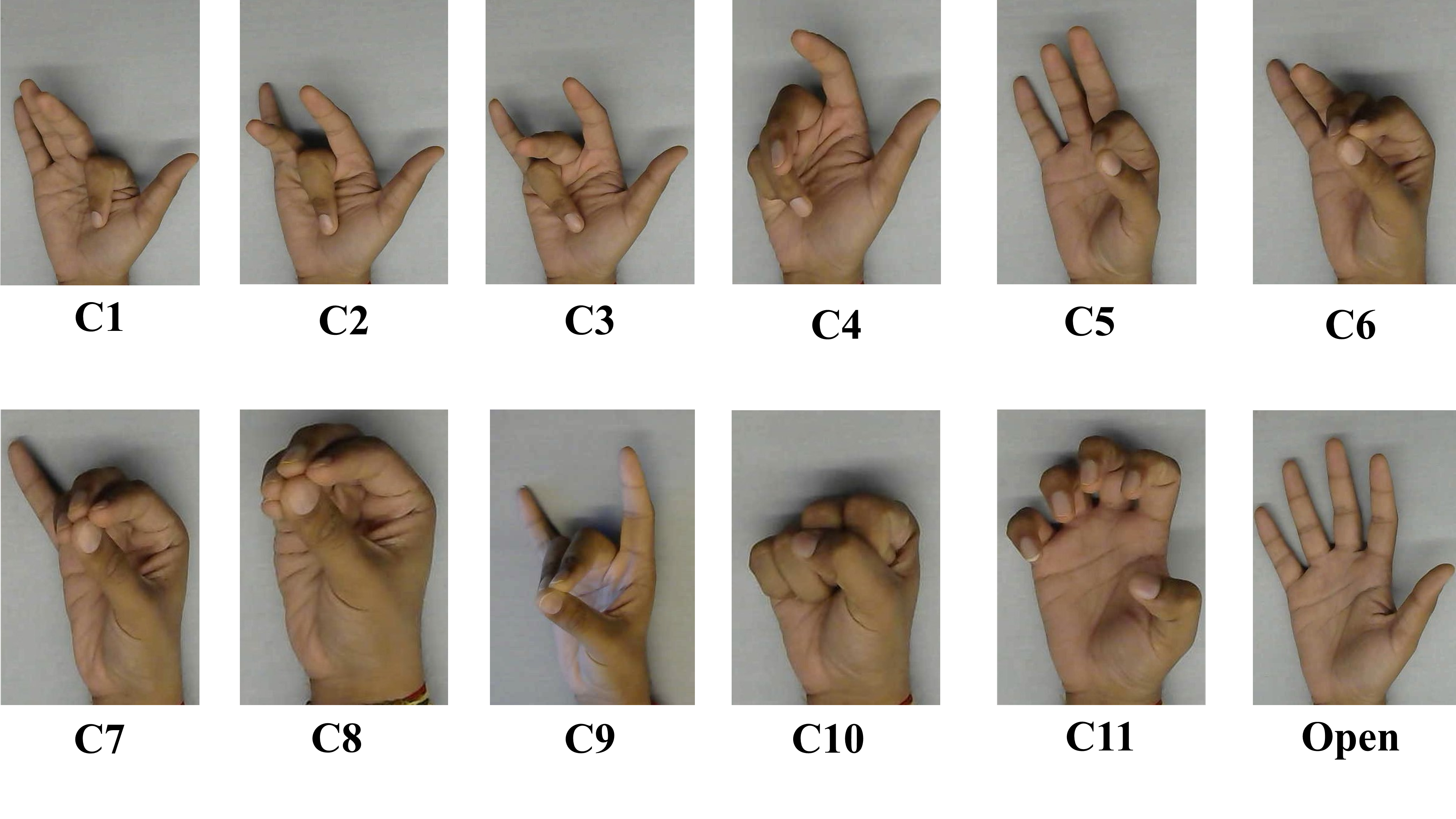}
\caption{The hand configurations: C1 (IndFlex): Index finger flexion; C2 (MidFlex): Middle finger flexion; C3 (RinFlex): Ring finger flexion; C4 (PinFlex): Pinky finger flexion; C5 (IndPinch): Index finger in contact with the thumb; C6 (IndMidPinch): Index \& Middle fingers in contact with the thumb; C7 (IndMidRinPinch): Index, Middle \& Ring fingers in contact with the thumb; C8 (AllPinch): All fingertips touching; C9 (MidRinPinch): Middle \& Ring fingers in contact with the thumb; C10 (Fist); C11 (Hook): Movement is restrained to just the interphalangeal joint movement with the MCP joint angle remaining constant; Open: All fingers extended.}
\label{fig:states}
\end{figure}
\subsubsection{Synchronization of Vicon and Verasonics systems}
Because of a sample rate mismatch between the trigger from the Verasonics ultrasound system and the Vicon motion capture system, an Arduino Uno was used as a bridge to record trigger signals utilizing Arduino's interrupt service routine and generate a pulse that could be read by the Vicon analog to digital converter.
\subsubsection{Audio Signal Generation}
Three audio waveforms were designed for simulating three different speeds for hand movements using Audacity, an open source digital audio workstation. 0.1 second Sawtooth waves at 440 Hz and 392 Hz were used to alert the subjects to alternate between the rest and motion hand states. The frequencies of the rest/motion switching were set as 2 Hz for fast, 1 Hz for medium \& 0.5 Hz for slow speeds. 
\subsubsection{Computation Hardware and Software}
NVIDIA GeForce RTX 2070 SUPER was used to train and run the angle prediction regression models. AMD Ryzen 7 2700X Eight-Core Processor was used to run and train the hand configuration classification models. The system had 31.91 GB available RAM. The code was executed in Python 3.7. TensorFlow Keras API was used to train and run deep learning models \cite{geron2019hands}. For SVC, scikit-learn library was used to run and train the machine learning based classification models \cite{pedregosa2011scikit}.

\subsection{Experimental Testing Protocol}
11 hand configurations relevant to the activities of daily living (ADLs) were chosen for this project \cite{dollar2014classifying}. The hand configurations are shown and described in the Fig \ref{fig:states}. The Open (all fingers extended) was considered as the rest state for all the 11 hand configurations and the subjects alternated between the open hand and selected the hand configuration. The subject was seated and 16 motion capture markers were attached to their right hand using a double sided tape. Motion capture markers were attached on approximate ends of metacarpal and proximal phalanx for each of the index, middle, ring and pinky fingers as seen in Fig. \ref{fig:Hand_Angle} (a). 

This marker configuration allowed us to calculate the MCP joint angle while ensuring free movement of the hand for the desired hand movements. The MCP joint angle was measured by taking the inverse cosine of the two vectors formed by each metacarpal-proximal phalanx pair. Ultrasound gel was applied to the subject's forearm and the imaging surface on the ultrasound probe. The ultrasound probe was then strapped to the subject's right forearm. Then, the subject's forearm was fastened to a rest table. The subject then performed a trial run for the 12 hand movements. Auditory beeps were used to indicate the desired rate of hand opening and closing. There was a rest of 1 minute between each data acquisition session and 2 minutes between each speed transition. 56 seconds of data was acquired for each data acquisition session. Per subject, there were 33 data acquisition sessions, for 11 different hand movements at 3 different speeds each.

\begin{figure*}[t]
\centering
\includegraphics[scale=0.17]{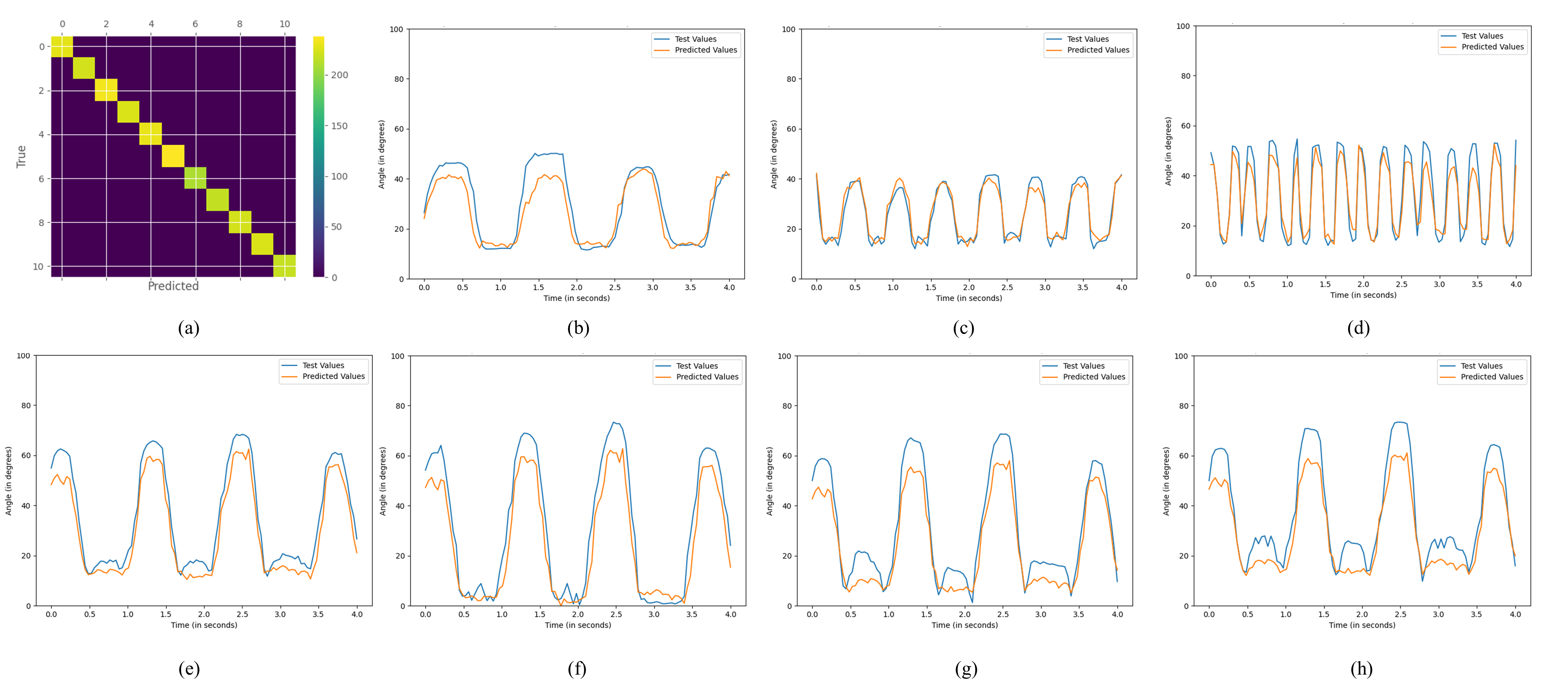}
\caption{(a) Confusion matrix for 11 state (C1 - C11: 0 - 10) classification for subject 1 with data combined for the three speeds using SVC. (b-d) For IndMidRinPinch, index finger predictions at slow (b), medium (c) and fast (d) speeds. (e-h) For fist hand configuration at medium speed, the index (e), middle (f), ring (g) and pinky (h) finger predictions. (b-h) are for data obtained from subject 1.}
\label{fig:cropped}
\end{figure*}
\section{Data Analysis}
Ultrasound data obtained was preprocessed before training the models by normalizing, log-compressing and reshaping to highlight relevant muscle features. For the 6 subjects we collected data for 11 hand configurations, each at 3 speeds, leading to a total of 198 sessions of data acquisition. Each file was was split into 30\% test data and 70\% training data.

\subsection{Data and Model Parameters} 
For each subject for each hand configuration and speed, 4 seconds  of  the training and 2 seconds of test data were saved. This data was then combined across the speeds for each subject and hand configuration leading to 12 seconds of training and 6 seconds of test data per hand configuration. All the individual training and testing files were combined and then were shuffled and split with a test-train split of 0.3 to minimize any bias. For our SVC implementation, we used a linear kernel because of the large number of features in each frame of the ultrasound image. We shuffled the data and set a test-train split of 0.3.

The CNNs were trained for each hand configuration and speed on 56 seconds of data for every subject. We used the Adam optimizer function as the gradient descent method \cite{kingma2014adam}. A learning rate of \(1e-3\) was set for the Adam optimizer. The learning rate decay was set to \(1e-3/200\). We used mean absolute error as the loss, which computes the mean of absolute difference between labels and predictions. Each data file was split with a test-train split of 0.3 without shuffling. While training, the validation split was defined as 0.1. The number of epochs was set to 50. 

\subsection{Angle Estimation for Ground Truth}
Motion capture markers were placed on the subject's hand at the base and the head of the proximal phalanges and the metacarpals for the index, middle, ring and pinky fingers. The marker locations on a subject's right hand can be seen in Fig \ref{fig:Hand_Angle}(a). Data from the thumb was omitted and will be the subject of future research. 

For the index metacarpal, the points can be defined as \(M1\) and \(M2\), with the vector between these points,  $\overrightarrow{M1 M2}$, shown in red in Fig \ref{fig:Hand_Angle}(a). For the index proximal phalanx, the points can be defined as \(P1\) and \(P2\), with the vector between these points,  $\overrightarrow{P1 P2}$, shown in blue in Fig \ref{fig:Hand_Angle}(a). These locations are based on the finger joints on the hand. By taking the dot product of the metacarpal and phalanges vectors extrapolated from the marker data, the MCP joint angles can be obtained.
In Fig \ref{fig:Hand_Angle}(b), the vectors and markers are shown for four different angles.

\subsection{Quantification Metrics}
For classification performance, classification accuracy percentage \(Acc\) was used as the evaluation metric. It is defined as
\[ Acc = \frac{TP + TN}{N} * 100\]

where, \(TP\) is the number of True Positives, \(TN\) is the number of True Negatives and \(N\) is the Total Sample Size. \(TP\) is an outcome where the model correctly predicts the positive class. \(TN\) is an outcome where the model correctly predicts the negative class. Root Mean Squared Error (\(RMSE\)) values were used to quantify the deep learning regression results. Root Mean Squared Error is defined as

\[RMSE = \sqrt{ \frac{1}{N}\sum_{i=1}^{N} (y_{i}-\hat{y_{i}})^2}\]

where, \(N\) is the Total Sample Size, \(i\) is the integer value ranging from 1 to the total number of samples, \(y_{i}\) is the test data value for the sample \(i\), and \(\hat{y_{i}}\) is the value predicted by the deep learning algorithm at the sample \(i\).

\section{Results}
Here we describe the hand configuration classification results using SVMs and MCP joint angle estimation results using CNNs. We also describe our successful low latency MCP joint angle prediction and hand configuration classification aimed at real time control of interfaces.

\subsection{Hand Configuration Classification}
11 classes described in Fig \ref{fig:states} were considered for the classification. We were able to obtain 100\% classification accuracy for an SVM classifier trained on all the speeds to reduce bias of our algorithm. Figure \ref{fig:cropped} (a) shows the confusion matrix of the classifier for subject \#1.

\subsection{Angle Estimation}
MCP joint angle prediction was done using VGG16, a CNN network used for to obtain relevant information from image data. Utilizing VGG16, we were able to generate a good degree of visual correspondence between our test data and the model predictions. This is shown for four second test data in Fig \ref{fig:cropped} (b) through Fig \ref{fig:cropped} (h). RMSE values were obtained from each of the hand configurations for each speed, for every subject. For the results, the range is set from 0\textdegree-100\textdegree to encapsulate the angle data from all subjects. The overall RMSE value averaged over all the hand configurations, three different speeds and all the subjects was found to be of 7.35\textdegree.
\subsubsection{Viability for different hand configurations}
RMSE values were calculated for the predictions averaged over all the subjects for all the slow, medium and fast speeds for each hand configuration. Fig \ref{fig:RMSE_Configs} shows the plot of RMSE results for different hand configurations. The lowest average RMSE value (4.55\textdegree) was obtained C11 (Hook) and the largest average RMSE value (11.43\textdegree) was obtained for C10 (Fist). 
\subsubsection{Repeatability over subjects}
\begin{figure}[t]
\centering
\includegraphics[scale=0.53]{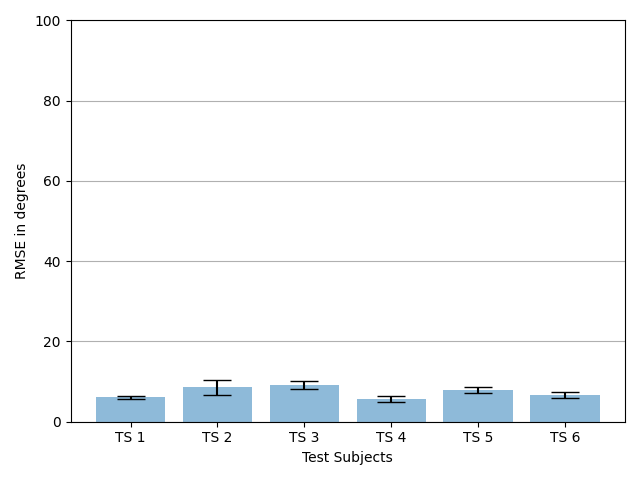}
\caption{RMSE results for different subjects averaged over the 3 speeds and different configurations.}
\label{fig:RMSE_TS}
\end{figure}

RMSE values were calculated for the predictions averaged over all the slow, medium and fast speeds, and the hand configuration for each subject. Fig \ref{fig:RMSE_TS} shows the plot of RMSE results for different subjects. The lowest average RMSE value (5.65\textdegree) was obtained from subject 4 and the largest average RMSE value (9.19\textdegree) was obtained from subject 3.

\subsection{Combined Prediction Pipeline}
For the combined prediction pipeline SVC models trained to classify between 11 different hand configurations for a single speed and subject and 11 VGG16 models trained for prediction of MCP joint angles were used. In the combined loop, first the hand configuration state prediction is done using the saved SVC model. Then, based on the SVC prediction, the CNN model is chosen from a dictionary of saved models. For each frame of data, the SVC prediction takes 0.1 - 0.15 seconds. CNN prediction takes 0.004 - 0.007 seconds. The total time ranges from 0.11 - 0.16 seconds for angle and state prediction for one frame of ultrasound information, leading to 6.25 - 9.1 Hz of data processing capability using the combined prediction pipeline.

\begin{figure}[t]
\centering
\includegraphics[scale=0.53]{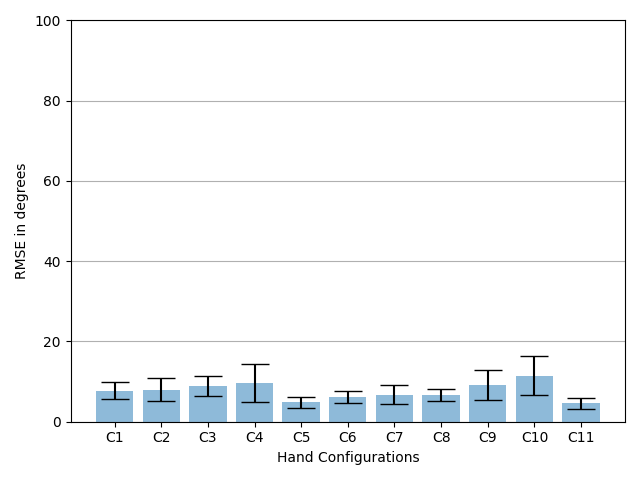}
\caption{RMSE results for hand configurations averaged over the 3 speeds.}
\label{fig:RMSE_Configs}
\end{figure}

\section{Discussion and Future Work}
Our experiments show that it is possible to obtain both hand configuration classification as well as finger angle estimation using one frame of ultrasound data using different classification and regression models. While ultrasound hand configuration classification is a well researched task, we were able to show that with our experimental setup we were able to obtain 100\% accuracy on ultrasound data combined for different speeds. These results were consistent with the high classification accuracy obtained by McIntosh et al. \cite{mcintosh2017echoflex}. Due to computational limitations, we were not able to train classifiers which were based on ultrasound data which was more than 198 seconds in length. It would be worthwhile to check out the classifier performance for bigger data-sets and to see how deep learning based classification performs on ultrasound data in comparison with the traditional SVC. For real time control applications, it would be necessary to evaluate the classification in an on-the-fly training and evaluation paradigm, similar to the 4 state discretized classifier trained in \cite{bimbraw2020towards}. We trained our SVC model on data acquired in a way that motion artifacts are minimised, and the subjects were supervised in a way that they would be consistent with their movements for a particular hand configuration movement set which lead to obtaining 100\% classification accuracy on our combined speed data.

During the data acquisition, because of erratic hand movements and motion capture marker occlusion, there were issues with getting proper ground truth for some files. Because of these issues, we were only able to get the RMSE results for 70\% of the total data files that we initially acquired. As can be seen in Fig \ref{fig:RMSE_Configs}, configurations C4 and C10 have the biggest RMSE values. This is because both these states (PinFlex and Fist) have significant pinky finger movement. Pinky finger movements were not consistent because it's hard to have repeatable movements with pinky finger. For configuration C11 (Hook), the RMSE value was the minimum because the MCP angle values are smaller than other hand configurations. 

For the combined prediction pipeline, most of the time is taken by SVC prediction since the current implementation is not optimized on GPU and it uses the system CPU which is much slower. We believe that exploring more GPU supported deep neural network based classification algorithms can significantly improve the rate of simultaneous prediction from the current 6.25 - 9.1 Hz. In the current paper, different deep learning models were trained for each angle estimation data. It would be worthwhile to explore deep learning algorithms which can generalise the MCP joint angle prediction for different hand configurations as well as for different subjects.

\section{Conclusions}
In this paper, we shared our results for MCP joint angle estimation and hand state prediction based on ultrasound data acquired from the forearm. Our CNN based model was able to get an RMSE with an average value of  7.35\textdegree \thinspace over 6 subjects and 11 hand configurations. Despite our limited compute for SVC, we were able to get promising classification results. We presented our combined pipeline for angle and hand state estimation which, on our current setup can generate predictions as fast as 6.25 - 9.1 Hz per frame of ultrasound data. These results are encouraging and they show the capability of ultrasound in understanding muscle movements. We hope that this work can inspire research in the promising domain of utilizing ultrasound for predicting both continuous and discrete hand movements which can be useful for an intuitive and adaptable control of physical robots and non-physical digital and AR/VR interfaces.

\bibliography{root.bib}

\begin{thebibliography}{10}

\bibitem{tarasenko2020artificial}
Anna Tarasenko, Mikheil Oganesyan, Daryna Ivaskevych, Sergii Tukaiev, Dauren
  Toleukhanov, and Nickolai Vysokov.
\newblock Artificial intelligence, brains, and beyond: Imperial college london
  neurotechnology symposium, 2020.
\newblock {\em Bioelectricity}, 2(3):310--313, 2020.

\bibitem{cheok2019review}
Ming~Jin Cheok, Zaid Omar, and Mohamed~Hisham Jaward.
\newblock A review of hand gesture and sign language recognition techniques.
\newblock {\em International Journal of Machine Learning and Cybernetics},
  10(1):131--153, 2019.

\bibitem{ahsan2009emg}
Md~Rezwanul Ahsan, Muhammad~I Ibrahimy, Othman~O Khalifa, et~al.
\newblock Emg signal classification for human computer interaction: a review.
\newblock {\em European Journal of Scientific Research}, 33(3):480--501, 2009.

\bibitem{jiang2017exploration}
Xianta Jiang, Lukas-Karim Merhi, Zhen~Gang Xiao, and Carlo Menon.
\newblock Exploration of force myography and surface electromyography in hand
  gesture classification.
\newblock {\em Medical engineering \& physics}, 41:63--73, 2017.

\bibitem{oudah2020hand}
Munir Oudah, Ali Al-Naji, and Javaan Chahl.
\newblock Hand gesture recognition based on computer vision: a review of
  techniques.
\newblock {\em journal of Imaging}, 6(8):73, 2020.

\bibitem{guo2021human}
Lin Guo, Zongxing Lu, and Ligang Yao.
\newblock Human-machine interaction sensing technology based on hand gesture
  recognition: A review.
\newblock {\em IEEE Transactions on Human-Machine Systems}, 2021.

\bibitem{suarez2012hand}
Jesus Suarez and Robin~R Murphy.
\newblock Hand gesture recognition with depth images: A review.
\newblock In {\em 2012 IEEE RO-MAN: the 21st IEEE international symposium on
  robot and human interactive communication}, pages 411--417. IEEE, 2012.

\bibitem{ahmed2020device}
Hasmath Farhana~Thariq Ahmed, Hafisoh Ahmad, and CV~Aravind.
\newblock Device free human gesture recognition using wi-fi csi: A survey.
\newblock {\em Engineering Applications of Artificial Intelligence}, 87:103281,
  2020.

\bibitem{li2020review}
Kexiang Li, Jianhua Zhang, Lingfeng Wang, Minglu Zhang, Jiayi Li, and Shancheng
  Bao.
\newblock A review of the key technologies for semg-based human-robot
  interaction systems.
\newblock {\em Biomedical Signal Processing and Control}, 62:102074, 2020.

\bibitem{graupe1982multifunctional}
Daniel Graupe, Javad Salahi, and Kate~H Kohn.
\newblock Multifunctional prosthesis and orthosis control via microcomputer
  identification of temporal pattern differences in single-site myoelectric
  signals.
\newblock {\em Journal of Biomedical Engineering}, 4(1):17--22, 1982.

\bibitem{lee2020image}
Ung~Hee Lee, Justin Bi, Rishi Patel, David Fouhey, and Elliott Rouse.
\newblock Image transformation and cnns: A strategy for encoding human
  locomotor intent for autonomous wearable robots.
\newblock {\em IEEE Robotics and Automation Letters}, 5(4):5440--5447, 2020.

\bibitem{kahanowich2021robust}
Nadav~D Kahanowich and Avishai Sintov.
\newblock Robust classification of grasped objects in intuitive human-robot
  collaboration using a wearable force-myography device.
\newblock {\em IEEE Robotics and Automation Letters}, 6(2):1192--1199, 2021.

\bibitem{dong2021soft}
Wentao Dong, Lin Yang, Raffaele Gravina, and Giancarlo Fortino.
\newblock Soft wrist-worn multi-functional sensor array for real-time hand
  gesture recognition.
\newblock {\em IEEE Sensors Journal}, 2021.

\bibitem{rabe2020use}
Kaitlin~G Rabe, Mohammad~Hassan Jahanandish, Kenneth Hoyt, and Nicholas~P Fey.
\newblock Use of sonomyographic sensing to estimate knee angular velocity
  during varying modes of ambulation.
\newblock In {\em 2020 42nd Annual International Conference of the IEEE
  Engineering in Medicine \& Biology Society (EMBC)}, pages 3799--3802. IEEE,
  2020.

\bibitem{araki2012artificial}
Nozomu Araki, Kenji Inaya, Yasuo Konishi, and Kunihiko Mabuchi.
\newblock An artificial finger robot motion control based on finger joint angle
  estimation from emg signals for a robot prosthetic hand system.
\newblock In {\em The 2012 international conference on advanced mechatronic
  systems}, pages 109--111. IEEE, 2012.

\bibitem{shrirao2009neural}
Nikhil~A Shrirao, Narender~P Reddy, and Durga~R Kosuri.
\newblock Neural network committees for finger joint angle estimation from
  surface emg signals.
\newblock {\em Biomedical engineering online}, 8(1):1--11, 2009.

\bibitem{wang2020semg}
Chao Wang, Weiyu Guo, Hang Zhang, Linlin Guo, Changcheng Huang, and Chuang Lin.
\newblock semg-based continuous estimation of grasp movements by long-short
  term memory network.
\newblock {\em Biomedical Signal Processing and Control}, 59:101774, 2020.

\bibitem{xiong2021deep}
Dezhen Xiong, Daohui Zhang, Xingang Zhao, and Yiwen Zhao.
\newblock Deep learning for emg-based human-machine interaction: A review.
\newblock {\em IEEE/CAA Journal of Automatica Sinica}, 8(3):512--533, 2021.

\bibitem{bimbraw2020towards}
Keshav Bimbraw, Elizabeth Fox, Gil Weinberg, and Frank~L Hammond.
\newblock Towards sonomyography-based real-time control of powered prosthesis
  grasp synergies.
\newblock In {\em 2020 42nd Annual International Conference of the IEEE
  Engineering in Medicine \& Biology Society (EMBC)}, pages 4753--4757. IEEE,
  2020.

\bibitem{zheng2006sonomyography}
Yong-Ping Zheng, MMF Chan, Jun Shi, Xin Chen, and Qing-Hua Huang.
\newblock Sonomyography: Monitoring morphological changes of forearm muscles in
  actions with the feasibility for the control of powered prosthesis.
\newblock {\em Medical engineering \& physics}, 28(5):405--415, 2006.

\bibitem{chen2010sonomyography}
Xin Chen, Yong-Ping Zheng, Jing-Yi Guo, and Jun Shi.
\newblock Sonomyography (smg) control for powered prosthetic hand: a study with
  normal subjects.
\newblock {\em Ultrasound in medicine \& biology}, 36(7):1076--1088, 2010.

\bibitem{shi2010feasibility}
Jun Shi, Qian Chang, Yong-Ping Zheng, et~al.
\newblock Feasibility of controlling prosthetic hand using sonomyography signal
  in real time: preliminary study.
\newblock 2010.

\bibitem{akhlaghi2015real}
Nima Akhlaghi, Clayton~A Baker, Mohamed Lahlou, Hozaifah Zafar, Karthik~G
  Murthy, Huzefa~S Rangwala, Jana Kosecka, Wilsaan~M Joiner, Joseph~J
  Pancrazio, and Siddhartha Sikdar.
\newblock Real-time classification of hand motions using ultrasound imaging of
  forearm muscles.
\newblock {\em IEEE Transactions on Biomedical Engineering}, 63(8):1687--1698,
  2015.

\bibitem{yang2020simultaneous}
Xingchen Yang, Jipeng Yan, Yinfeng Fang, Dalin Zhou, and Honghai Liu.
\newblock Simultaneous prediction of wrist/hand motion via wearable ultrasound
  sensing.
\newblock {\em IEEE Transactions on Neural Systems and Rehabilitation
  Engineering}, 28(4):970--977, 2020.

\bibitem{mcintosh2017echoflex}
Jess McIntosh, Asier Marzo, Mike Fraser, and Carol Phillips.
\newblock Echoflex: Hand gesture recognition using ultrasound imaging.
\newblock In {\em Proceedings of the 2017 CHI Conference on Human Factors in
  Computing Systems}, pages 1923--1934, 2017.

\bibitem{simonyan2014very}
Karen Simonyan and Andrew Zisserman.
\newblock Very deep convolutional networks for large-scale image recognition.
\newblock {\em arXiv preprint arXiv:1409.1556}, 2014.

\bibitem{geron2019hands}
Aur{\'e}lien G{\'e}ron.
\newblock {\em Hands-on machine learning with Scikit-Learn, Keras, and
  TensorFlow: Concepts, tools, and techniques to build intelligent systems}.
\newblock O'Reilly Media, 2019.

\bibitem{pedregosa2011scikit}
Fabian Pedregosa, Ga{\"e}l Varoquaux, Alexandre Gramfort, Vincent Michel,
  Bertrand Thirion, Olivier Grisel, Mathieu Blondel, Peter Prettenhofer, Ron
  Weiss, Vincent Dubourg, et~al.
\newblock Scikit-learn: Machine learning in python.
\newblock {\em the Journal of machine Learning research}, 12:2825--2830, 2011.

\bibitem{dollar2014classifying}
Aaron~M Dollar.
\newblock Classifying human hand use and the activities of daily living.
\newblock In {\em The Human Hand as an Inspiration for Robot Hand Development},
  pages 201--216. Springer, 2014.

\bibitem{kingma2014adam}
Diederik~P Kingma and Jimmy Ba.
\newblock Adam: A method for stochastic optimization.
\newblock {\em arXiv preprint arXiv:1412.6980}, 2014.

\end{thebibliography}

\end{document}